\RequirePackage{snapshot}
\documentclass[10pt,twocolumn,letterpaper]{article}

\usepackage{times}
\usepackage{epsfig}
\usepackage{graphicx}
\usepackage{amsmath}
\usepackage{amssymb}

\usepackage{eso-pic}
\usepackage{xspace}

\def\abstract{\centerline{\large\bf Abstract}\vspace*{12pt}\it}

\usepackage{xspace}

\makeatletter
\DeclareRobustCommand\onedot{\futurelet\@let@token\@onedot}
\def\@onedot{\ifx\@let@token.\else.\null\fi\xspace}
\def\eg{\emph{e.g}\onedot}

\def\etal{\emph{et al}\onedot}
\makeatother

\usepackage{titlesec}
\usepackage{etoolbox}
\makeatletter
\patchcmd{\ttlh@hang}{\parindent\z@}{\parindent\z@\leavevmode}{}{}
\patchcmd{\ttlh@hang}{\noindent}{}{}{}
\makeatother

\titlespacing {\section}
{0pt}{1.2ex plus 1ex minus .2ex}{1.0ex plus .2ex}
\titlespacing {\subsection}
{0pt}{1.0ex plus 1ex minus .2ex}{1.0ex plus .2ex}
\titleformat{\section}
  {\normalfont\fontsize{12}{15}\bfseries}{\thesection.}{0.3em}{}
\titleformat{\subsection}
  {\normalfont\fontsize{11}{15}\bfseries}{\thesubsection}{0.3em}{}

\usepackage{caption}
\usepackage{gensymb} 
\usepackage{textcomp} 
\usepackage{enumitem}
\setitemize{noitemsep,topsep=0pt,parsep=0pt,partopsep=0pt}

\usepackage[pagebackref=true,breaklinks=true,letterpaper=true,colorlinks,bookmarksnumbered=true]{hyperref}
\hypersetup{
    pdftitle={Long Range 3D with Quadocular Thermal (LWIR) Camera},
    pdfauthor={Andrey Filippov, Oleg Dzhimiev},
    pdfkeywords={lwir, 3d},
    bookmarksnumbered=true,     
    bookmarksopen=true,         
    bookmarksopenlevel=1,       
    colorlinks=true,            
    pdfstartview=Fit,           
    pdfpagemode=UseOutlines,
    pdfpagelayout=TwoPageRight
}

\usepackage{enumitem}
\setitemize{itemsep=1pt,topsep=0pt,parsep=1pt,partopsep=0pt}
\setenumerate{itemsep=1pt,topsep=0pt,parsep=1pt,partopsep=0pt}

\usepackage{textgreek} 
\usepackage{siunitx}
\usepackage[T1]{fontenc}
\DeclareSIUnit[number-unit-product = {}]{\inchQ}{\textquotedbl}
\DeclareSIUnit[number-unit-product = {\thinspace}]{\inch}{in}
\DeclareSIUnit[number-unit-product = {}]{\pixel}{pix}
\usepackage[shortcuts,nopostdot,toc,nogroupskip,section,numberedsection=autolabel]{glossaries}
\usepackage{xcolor}
\definecolor{glscolor}{rgb}{0.23, 0.27, 0.29}
\setkeys{glslink}{hyper=false}
\renewcommand*{\CustomAcronymFields}{%
  name={\the\glsshorttok},%
  description={\the\glslongtok},%
}

\SetCustomStyle

\makeglossaries 
\newacronym{uas}{UAS}{unmanned aerial system}
\newacronym{uav}{UAV}{unmanned aerial vehicle}
\newacronym{suas}{sUAS}{small unmanned aerial system}
\newacronym{suass}{sUASS}{small unmanned aerial system swarm}
\newacronym
[description=counter-\gls{suass} system]
{csuass}{C-sUASS}
{Counter-sUASS}

\newacronym{lmfc}{LMFC}{Lockheed Martin Missile and Fire Control}
\newacronym{swap} {SWaP}{size, weight and power}
\newacronym{swapc}{SWaP-C}{size, weight, power, and cost}

\newacronym
[description=Light Detection And Ranging{,} originally ``light radar'']
{lidar}{LIDAR}{Light Detection And Ranging}

\newacronym
[description=high power continuous and pulsed laser and microwave weapon
systems] {dew}{DEW}{Directed Energy Weapons}

\newacronym{gps}{GPS}{Global Positioning System}
\newacronym{helmtt}{HELMTT}{High Energy Laser Mobile Test Truck}
\newacronym{mehel}{MEHEL}{Mobile Expeditionary High-Energy Laser}

\newacronym[description={short wave infrared (1.4-3.0~\textmu m)}]
{swir}{SWIR}{Short Wave Infrared}
\newacronym[description={middle wave infrared (3-8~\textmu m)}]
{mwir}{MWIR}{Middle Wave Infrared}
\newacronym[description={long wave infrared (8-15~\textmu m)}]
{lwir}{LWIR}{Long Wave Infrared}

\newacronym
[description={visible and near infrared, light bandwidth detected by ordinary
image sensors}]
{vnir}{VNIR}{Visible and Near Infrared}

\newacronym
[description={electro-optical (visible range) imaging system}]
{eo}{EO}{Elecro-optical}

\newacronym
[description={electro-optical (visible range) and infrared imaging system}]
{eoir}{EO/IR}{Elecro-optical and Infrared}

\newacronym{adsb}{ADS-B}{{automatic dependent surveillance -- broadcast}}
\newacronym{spad}{SPAD}{single-photon avalanche diode}
\newacronym{fov}{FoV}{field of view}
\newacronym{hfov}{HFoV}{horizontal field of view}
\newacronym{roi}{RoI}{region of interest}
\newacronym{tof}{ToF}{time-of-flight}
\newacronym{rf}{RF}{radio frequency}
\newacronym{dnn}{DNN}{deep neural network}
\newacronym
[description={image sensor with the circuit board, lens and optional shutter}]
{sfe}{SFE}{sensor front end}

\newacronym
[description={central processing unit, main general purpose processor of a
system}] {cpu}{CPU}{central processing unit}

\newacronym
[description={graphic processing unit, used for  
massive parallel processing, including \gls{ml}}] {gpu}{GPU}{graphic
processing unit}

\newacronym{asic}{ASIC}{application-specific integrated circuit}

\newacronym{soc}{SoC}{system-on-chip}

\newacronym
[description={vision processing unit - \gls{asic} dedicated to image/video
processing, \gls{ml} acceleration}] {vpu}{VPU}{vision processing unit}

\newacronym
[description={Tile Processor - Elphel software/\gls{rtl} for 
transform-domain image processing}] {tp}{TP}{Tile Processor}

\newacronym{cots}{COTS}{commercial off-the-shelf}

\newacronym
[description={focal plane array - image sensor, including visible, infrared
and any other wavelength}] {fpa}{FPA}{focal plane array}

\newacronym{piv}{PIV}{Particle Image Velocimetry}
\newacronym{psf}{PSF}{point spread function}
\newacronym{cad}{CAD}{computer aided design}
\newacronym{sn}{S/N}{signal-to-noise}

\newacronym
[description={Electronic Rolling Shutter - image sensor exposure method, where
pixels are exposed in line-scan order (opposite to snapshot shutter)}]
{ers}{ERS}{electronic rolling shutter}

\newacronym
[description={Frequency Domain, usually same as Transform Domain - data
transformed by Fourier or similar transform, enables efficient
implementation of convolution and correlation}]
{fd}{FD}{Frequency Domain}

\newacronym{dct}{DCT}{discrete cosine transform}
\newacronym{dst}{DST}{discrete sine transform}
\newacronym
[description={Discrete Trigonometric Transform - common name for shared by 
\gls{dct} and \gls{dst}}]
{dtt}{DTT}{discrete trigonometric transform}

\newacronym
[description={Modified Complex Lapped Transform - invertible transform
to/from \gls{fd} based on \gls{dct} and \gls{dst} type~IV}] {mclt}{MCLT
}{modified complex lapped transform }

\newacronym{dofr}{DoF}{degrees of freedom}

\newacronym{tdac}{TDAC}{time-domain aliasing cancellation}

\newacronym{ssd}{SSD}{solid state drive}

\newglossaryentry{jp4}{name={JP4}, description={image format
developed by Elphel for efficient compression of raw Bayer mosaic images; first
used for Google Books project}}

\newacronym{rmse}{RMSE}{root mean square error}

\newacronym
[description={register-transfer level - design abstraction level in
hardware description languages}] {rtl}{RTL}{register-transfer level}

\newacronym{floss}{FLOSS}{free, libre, and open source software}
\newacronym{fpga}{FPGA}{field-programmable gate array}
\newacronym{pld}{PLD}{programmable logic devices}

\newacronym{ml}{ML}{machine learning}
\newacronym{lma}{LMA}{Levenberg-Marquardt algorithm}

\newglossaryentry{tpnet}{name={TPNET}, description={\gls{tp} with \gls{dnn}
for 3D perception system}}

\newacronym
[description={Google \gls{asic} for \gls{ml} acceleration}]
{tpu}{TPU}{tensor processing unit}
\newacronym{ai}{AI}{artificial intelligence}
\newacronym{ugv}{UGV}{unmanned ground vehicle}
\newacronym{agv}{AGV}{autonomous ground vehicle}
\newacronym{slam}{SLAM}{simultaneous localization and mapping}

\newglossaryentry{sonar}{name={sonar}, description={ultrasound ranging  device,
originally acronym for SOund NAvigation Ranging}}

\newacronym{ar}{AR}{augmented reality}
\newacronym{pcb}{PCB}{printed circuit board}
\newacronym{ara}{ARA}{Applied Research Associates}
\newacronym{poc-corp}{POC}{Physical Optics Corporation}
\newacronym{ims}{IMS}{inertial measurement system}
\newacronym{aas}{AAS}{autonomous amphibious system}
\newglossaryentry{lm-athena}{name={ATHENA}, description={Lockheed
Martin Advanced Test High Energy Asset}}
\newglossaryentry{ray-phaser}{name={PHASER}, description={Raytheon
anti-drone high-powered microwave emitter}}

\newglossaryentry{gnugpl}{name={GNU/GPL}, description={GNU General Public
License}}
\newglossaryentry{cernohl}{name={CERN OHL}, description={CERN (European
Organization for Nuclear Research) Open Hardware License}}
\newacronym{atak}{ATAK}{Android Team Awareness Kit}
\newacronym{afrl}{AFRL}{Air Force Research Laboratory}
\newacronym{anit}{ANIT}{alternative night/day imaging technologies}
\newacronym{ivas}{IVAS}{U.S. Army Futures Command Integrated Visual Augmentation
System}
\newacronym{hud3}{HUD 3.0}{Heads Up Display 3.0}
\newacronym{fps}{FPS}{frames per second}
\newacronym{sbir-ca}{CA}{contract award}
\newacronym{rv}{RV}{reentry vehicle}
\newacronym{sbir}{SBIR}{Small Business Innovation Research}
\newacronym{em}{EM}{electromagnetic}
\newacronym{sfm}{SfM}{Structure-from-Motion}
\newacronym{hgv}{HGV}{Hypersonic Glide Vehicle}
\newacronym{ssl-mda}{SSL}{Space Sensor Layer}
\newacronym{hbtss}{HBTSS}{Hypersonic and Ballistic Tracking Space Sensor}
\newacronym{pleo}{P-LEO}{Proliferated Low Earth Orbit}
\newacronym{mre}{MRE}{mean reprojection error}
\newacronym{ti}{TI}{thermal imaging}
\newacronym
[description={Stereo images rectification to the common for the set distortion
model}] {difrec}{DR}{differential rectification}
\newacronym{rsd}{RSD}{relative standard deviation}
\newacronym{svd}{SVD}{singular value decomposition}
\newacronym{fc}{FC}{fully connected}
\newacronym{fg}{FG}{foreground}
\newacronym{bg}{BG}{background}

\begin{document}

\title{Long Range 3D with Quadocular Thermal (LWIR) Camera}

\author{
Andrey Filippov \qquad Oleg Dzhimiev \\
Elphel, Inc. 1455 W. 2200 S. \#205, Salt Lake City, Utah 84119 USA \\
{\tt\small \{andrey,oleg\}@elphel.com}
}

\maketitle

\glsunset{lidar}

\begin{abstract}
   \Gls{lwir} cameras provide images regardless of the ambient illumination, they
   tolerate fog and are not blinded by the incoming car headlights. These
   features make \gls{lwir} cameras attractive for autonomous navigation,
   security and military applications. Thermal images can be used similarly to
   the visible range ones, including 3D scene reconstruction with two or more
   such cameras mounted on a rigid frame. There are two additional challenges
   for this spectral range: lower image resolution and lower contrast of
   the textures. 
   
   In this work, we demonstrate quadocular \gls{lwir} camera setup, calibration,
   image capturing and processing that result in long range 3D perception with
   0.077~pix disparity error over 90\% of the depth map. With low resolution
   ($160\times 120$) \gls{lwir} sensors we achieved 10\% range accuracy at 28~m
   with $56^{\circ}$ \gls{hfov} and 150~mm baseline. Scaled to the now-standard
   $640\times 512$ resolution and 200~mm baseline suitable for
   head-mounted application the result would be 10\% accuracy at 130~m.
\end{abstract}


\section{Introduction}
\label{sec:introduction}

\begin{figure}[t]
\begin{center}
\includegraphics[width=1.0\linewidth]{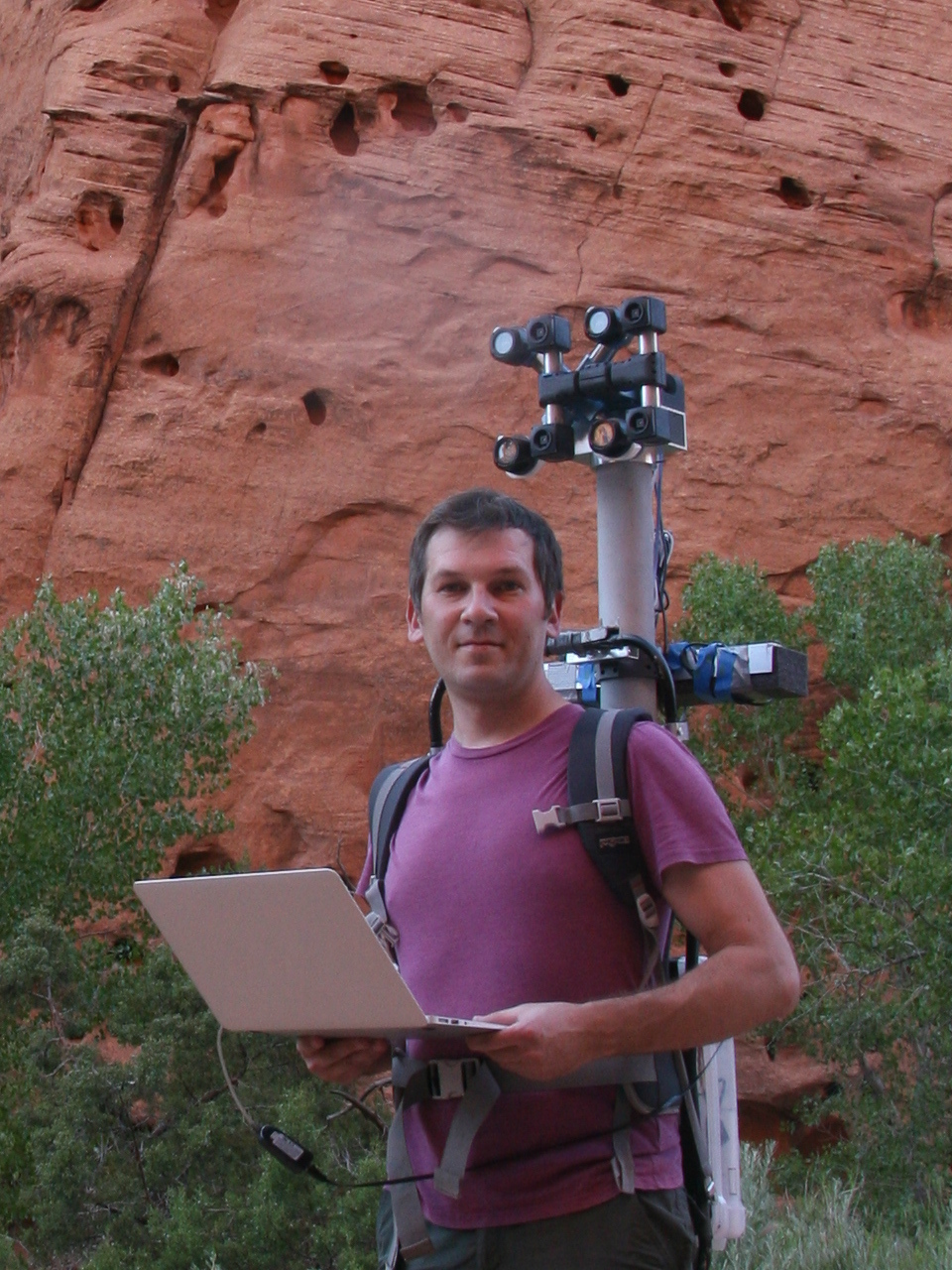}
\end{center}
\caption{Dual quadocular \gls{lwir}+RGB experimental camera.}
\label{fig:oleg_talon}
\end{figure}

Advances in uncooled \gls{lwir} detectors based on microbolometer arrays made
\gls{ti} practical in multiple application areas where remote object's temperature
is important itself or where the temperature gradients and differences can be
used for localization, object detection, and tracking, 3D scene reconstruction.
Temperature measurements combined with 3D data are needed in medical
applications (Chernov~\etal~\cite{chernov20173d},
Cao~\etal~\cite{cao2018depth}), energy auditing (Vidas and
Moghadam~\cite{vidas2013heatwave}, \cite{cao2018depth}). High thermal contrast
of live objects facilitates \gls{fg}/\gls{bg} separation in 3D models for
counting pedestrians (Kristoffersen~\etal~\cite{kristoffersen2016pedestrian}).
The main application of the \gls{ti} fused with the other data sources remains
autonomous and assisted driving. While not considering specifically \gls{ti} 3D,
Miethig~\etal~\cite{miethig2019leveraging} compare it with other sensors
(\glspl{lidar}, radars, sonars, and RGB camera) arguing that the use of \gls{ti}
could have prevented some fatalities that were caused by autonomously driven cars.
They mention high contrast for the living organisms in any weather conditions,
resilience to fog, as well as the fact that \gls{lwir} sensors are not
saturated by the headlights of the incoming traffic.

Most of the known 3D perception systems rely on multimodal data, such as RGB,
\gls{ti}, and depth. Treible~\etal~\cite{treible2017cats} created a 3-modal
benchmark set, where synchronized binocular RGB and \gls{lwir} are
combined with a ground truth image captured with a \gls{lidar}. \Gls{lwir}-only 3D perception is
beneficial for the military applications where active ranging methods
(\gls{lidar}, \gls{tof} or structured light) are easily detectable by the
properly equipped adversary. 3D perception for \gls{ugv} was researched by
Lee~\etal~\cite{lee2016lwir}, Zapf~\etal~\cite{zapf2017perception}, and while
the experimental rig was tri-modal, they evaluated \gls{lwir}-only performance
too while using \gls{lidar} data as a ground truth.

We developed and evaluated a quadocular \gls{lwir} 3D perception system
(Figure~\ref{fig:oleg_talon}) that is paired with another RGB quadocular camera
(Elphel MNC393-XCAM) as a source of the ground truth data for the \gls{dnn}
training and evaluation. \Gls{lwir} system uses four \SI{160 x 120}{} FLIR
Lepton modules, the RGB cameras have \SI{2592 x
1936}{} sensors with $16\times$ higher linear resolution. This ratio and
almost identical values for the \gls{fov}, baseline, and \gls{mre}, measured in pixels
for both modalities made it possible to consider the depth map generated from
the RGB images as ground truth for the \gls{lwir} ones -- for every matched
object distance the expected range error for \gls{lwir} is 13 times larger than
that of the RGB modality.

\begin{figure}[t]
\begin{center}
\includegraphics[width=1.0\linewidth]{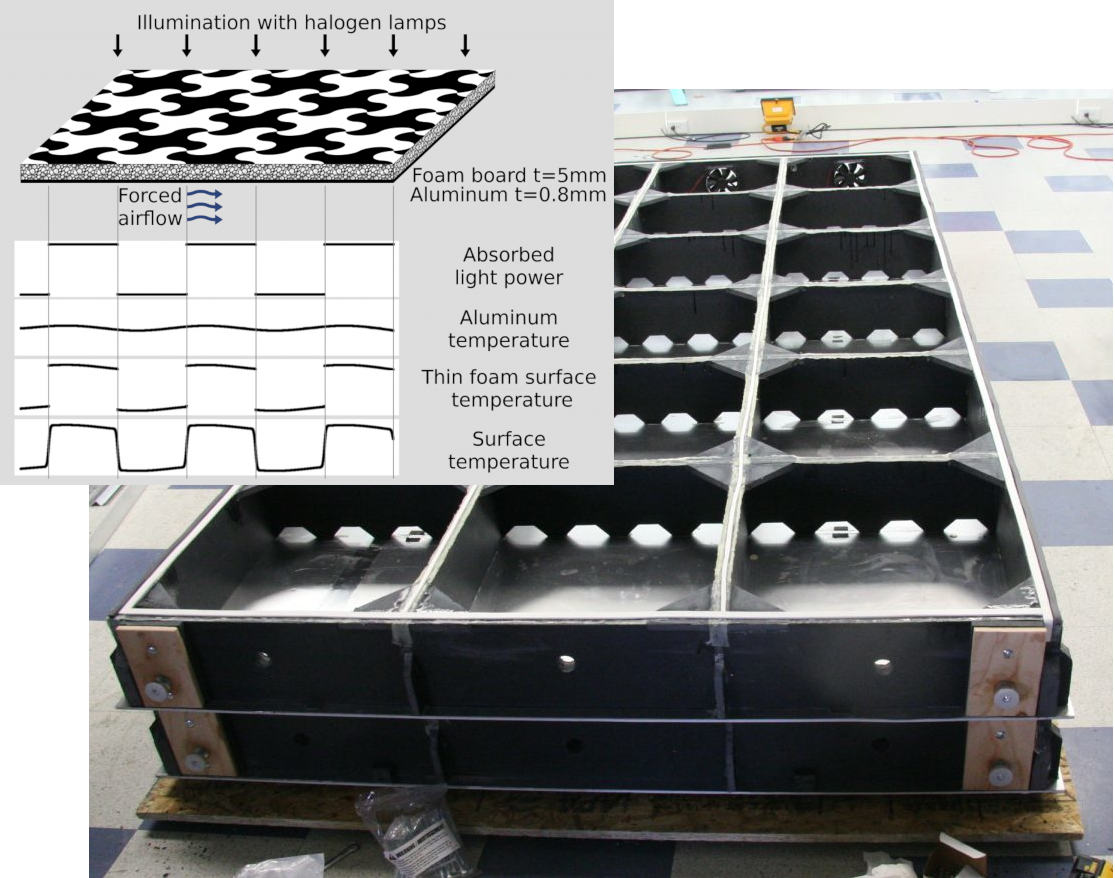}
\end{center}
\caption{\gls{lwir} calibration pattern design and a rear view of the calibration
panel.}
\label{fig:pattern_back}
\end{figure}

Our contributions are as follows:
\begin{enumerate}
  \item Development of a \gls{lwir} quadocular camera prototype for wearable and
  small \gls{ugv} applications;
  \item Development of the large format (\SI{7 x 3}{\meter}) dual-modal
  calibration system for high resolution photogrammetric and radiometric
  applications;
  \item Adaptation of the long range RGB image processing open source software (Filippov and
  Dzhimiev~\cite{filippov2018see}) to \gls{lwir}, evaluation of the prototype system and
  comparing its performance to state of the art.
\end{enumerate}
The rest of the paper is organized as follows. In Section~\ref{sec:calibration}
we overview existing methods of the \gls{lwir} cameras calibration and describe
our contribution. In Section~\ref{sec:acquisition} we explain our system
hardware, image sets acquisition, low-level preprocessing, and training of the
\gls{dnn} to improve depth map accuracy. Finally, in
Section~\ref{sec:discussion} we compare our results with the prior ones.

\section{Calibration}
\label{sec:calibration}

\begin{figure}[t]
\begin{center}
\includegraphics[width=1.0\linewidth]{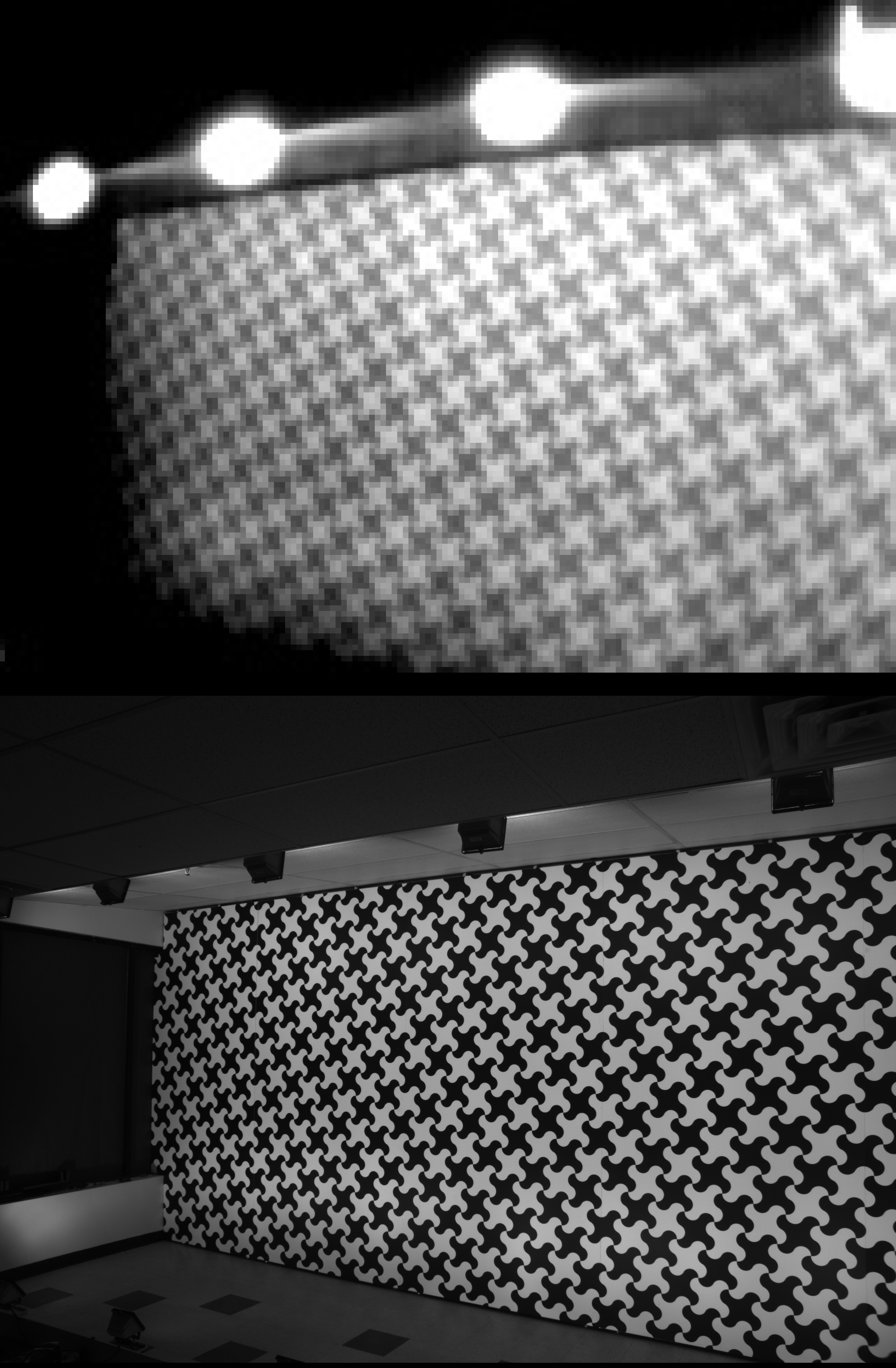}
\end{center}
\caption{Calibration pattern views: \gls{lwir} (top), visible range (bottom).}
\label{fig:pattern_views}
\end{figure}

\subsection{Existing \gls{lwir} and Multimodal Calibration}
\label{subsec:calibration_existing}

Photogrammetric camera calibration is an important precondition to achieve an
accurate depth map in the stereo application, and most methods use special
calibration setups to perform this task. It is convenient to use a standard
OpenCV calibration procedure, another popular (over 13,000 citations) approach
was presented by Zhang~\cite{zhang2000flexible}, it is used by many \gls{ti} and
multimodal calibration systems. This method involves capturing an image of a
checkerboard or a pattern of squares rotated at several angles (or moving the
camera around), matching the corners of the squares and then using \gls{lma} to simultaneously
find camera extrinsic and intrinsic  parameters by minimizing reprojection
error. The challenge of the \gls{lwir} and multimodal calibrations is to make
this pattern to be reliably registered by a thermal or all modalities.

Rangel~\etal~\cite{rangel20143d} provide a classification of the
\gls{ti} calibration for both active (heated by powered wires, resistors
or LEDs) and passive (printed or cut pattern heated by a floodlight or 
sun) designs. They tested several targets and selected cardboard with the
circular holes in front of a heated surface.
Vidas~\etal~\cite{vidas2012mask} used an A4-sized cardboard pattern with
\SI{20 x 20}{\milli\meter} holes and a computer monitor as a heated
\gls{bg}, reporting no blurring of the corners.
Skala~\etal~\cite{skala20114d} used Styrofoam panel with cut square holes
that worked well for both \gls{lwir} and depth sensors.

Just for the \gls{ti} alone Ng~\etal~\cite{ng2005acquisition} suggested a heated
wire net with plastic \gls{bg} on a turntable,
Prakash~\etal~\cite{prakash20063d} used printed checkerboard on paper with a
flood lamp.
Saponaro~\etal~\cite{saponaro2015improving} made the checkerboard corners
sharper by attaching the paper pattern to a ceramic tile, they also measured
the change of the target contrast over time after turning off the floodlight.
\Gls{lwir} image blur that makes matching points registration difficult was
targeted by several other works.
Harguess and Strange~\cite{harguess2014infrared} made a pattern with \SI{3 x
5}{} circles (\SI{170}{\milli\meter} pitch) printed on \SI{3}{\milli\meter}
aluminum composite Dibond consisting of two \SI{0.3}{\milli\meter} aluminum
layers with the polyethylene core between them.

St-Laurent~\etal~\cite{st2017passive} used a thermally conductive substrate
-- \SI{2.4}{\milli\meter} solid aluminum instead of the aluminum laminate
and achieved sharp \gls{lwir} images suitable for automatic corner detection
with OpenCV and used it with tri-modal \gls{lwir}, \gls{swir}, and visible
range. Similar to the previous work this pattern required high irradiance power
as the thermal difference was developed in the thin paint layer and
was usable only outdoors under the bright sunlight.
Rankin~\etal~\cite{rankin2011unmanned} evaluated multiple calibration targets
for \gls{lwir} and \gls{mwir}, passive ones (heated by the sunlight) had \SI{5 x
5}{} black dots printed on foamcore board attached to aluminum honeycomb, they
also used heated plastic inserts on a metal frame and a large \SI{1.3 x
1.5}{\meter} checkerboard with heated wires along the edges developed by General
Dynamics.

Other active methods depend on heated elements, such as light bulbs
(Yang~\etal~\cite{yang2011geometric},
Zoetgnand~\etal~\cite{zoetgnande2019robust}), resistors
(Gschwandtner~\etal~\cite{gschwandtner2011infrared}) or overheated
LEDs (Beauvisage and Aouf~\cite{beauvisage2017low},
Li~\etal~\cite{li2018spatial}).

Most of the referenced works report \gls{mre} in the range of few tenths of a
pixel (0.41~pix~\cite{saponaro2015improving},
0.2..0.8~pix~\cite{harguess2014infrared}, 0.1..0.2~pix~\cite{st2017passive},
0.22..0.25~pix~\cite{beauvisage2017low}, 0.36~pix~\cite{li2018spatial}, 0.3~pix,
~\cite{gschwandtner2011infrared}, 0.25~pix~\cite{zoetgnande2019robust}), but
Ellmauthaler~\etal~\cite{ellmauthaler2013novel} report much lower \gls{mre} or
0.036~pix and even 0.0287~pix in
Ellmauthaler~\etal~\cite{ellmauthaler2019visible}. While they are using an
iterative process that calculates centers of mass of the rectified image of
light bulbs instead of just coordinates of the corners, the reported
\gls{mre} may be too optimistic because the target pattern image in each frame
is too small.
We would expect significantly higher \gls{mre} if the pattern was
registered over the full frame.

Long range depth measurements require high disparity resolution, therefore,
calibration has to provide low \gls{mre} over the full \gls{fov}. We achieve
$\gls{mre}_{LWIR} = 0.067~pix, \gls{mre}_{RGB} = 0.036~pix$ by using a very
large calibration pattern, bundle-adjusting the pattern nodes 3D coordinates
simultaneously with calibrating of a multi-sensor high-resolution camera, and
use of the accurate phase correlation matching of the registered patterns.

\subsection{Dual-modality Calibration Target}
\label{subsec:calibration_pattern}

Our dual-modality visible/\gls{lwir} \SI{7.0 x 3.0}{\meter} calibration target
consists of five separate panels fitted together. The black-and-white pattern
resembles the checkerboard one with each straight edge replaced by a combination
of two arcs -- such layout makes spatial spectrum uniform compared to a plain
checkerboard pattern and facilitates \gls{psf} measurement of the cameras.

Figure~\ref{fig:pattern_back} shows the
construction and a back view of a \SI{1.5 x 3.0}{\meter} panel. This design
provides a definite correspondence between the image
registered by the visible range cameras and the surface temperature, resulting
in time-invariant images for both modalities.
The pattern illuminated by ten of \SI{500}{\watt} halogen floodlights is printed on a
\SI{5}{\milli\meter} foamcore board glued to an \SI{0.8}{\milli\meter} aluminum
sheet. We found that it is important to use both inks -- black and white 
because the unpainted paper surface of the white foamcore board behaves almost
like a mirror showing flood light reflection in \gls{lwir} spectral range, while the paint
bumps provide sufficient diffusion to mitigate this effect.

The pattern temperature does not depend much on uncontrolled convection
as the back side of the pattern is cooled by the forced airflow.
In our current experiments, we used the full power of the fans,
but it is easy to control the airflow in each target segment and precisely
maintain the temperature of the aluminum backing for radiometric calibration.
High thermal conductance of the aluminum and low conductance of the foamcore
board together with the small board thickness relative to the cell edge
(\SI{5}{\milli\meter} vs. \SI{180}{\milli\meter}) result in sharp and uniform
\gls{lwir} images.
For precise radiometric calibration, the residual
non-uniformity and blur can be modeled to calculate the
temperature from the light intensity registered by a visible range camera.
Each target segment is made mechanically rigid by the structural
elements cut from the thicker foamcore material. Additionally,
when calibration is in progress, the fans provide negative air pressure holding the
panels tightly against the wall.

During the calibration process several hundred image sets (4~RGB and
4~\gls{lwir}) are acquired using the robotic fixture, scanning
\SI{\pm 40}{\degree} horizontally and \SI{\pm 25}{\degree}
vertically with 80\% overlap from three locations: \SI{7.4}{meters} along the
pattern axis and then from the 2 (right and left) locations \SI{4.5}{\meter}
from the target and \SI{2.2}{\meter} from the center,
Figure~\ref{fig:pattern_views} shows example of the registered \gls{lwir} (top) and visible (bottom) images. The dark lower left of the
\gls{lwir} image is still processed -- dynamic range is limited only in this
preview.

While the accuracy of the camera rotation fixture is insufficient to output
camera pose for fitting, it resolves pattern ambiguity for
automatic matching of the images.

\subsection{Processing of the Calibration Image Sets}
\label{subsec:calibration_processing}

Acquired sets of 8 images and two camera rotation angles each are
processed with the open source Java code organized as a plugin to the popular
ImageJ framework in  three stages. The first one takes
hours to complete but is fully automatic. The program operates on
each image individually and generates coordinates of the pattern nodes that
correspond to the corners of the checkerboard pattern.

The pattern detection in the image is in turn handled in the following steps:
\begin {enumerate}
  \item Image is scanned in 2D reversed binary order (skipping areas where
  the pattern is already detected) for potential pattern fragments using
  \gls{fd} representation of patches containing 10-50 grid cells.
  \item If a potential match is found, the two grid vectors are determined, the
  corresponding synthetic grid patch is calculated, the phase correlation is
  found for the registered and synthetic grids, and if the result exceeds the
  specified threshold, then the cell is marked.
  \item Neighbor cells are searched by the wave algorithm around the newly found
  cells, each time calculating grid vectors from the known cells around. When
  the wave dies, the search is continued from Step 1 until exhausting all
  possible locations.
  \item The found pattern grid is refined by re-calculating phase correlation
  between the registered image and simulated patches for each detected node.
  This time the simulated pattern is built using second degree polynomials
  instead of just linear transformation used in the previous steps.
\end {enumerate}
For the convenience of development and visualization the produced data is saved
as a 4-slice 32-bpp Tiff file: X pixel coordinate as a float value, Y
coordinate, and the corresponding U, V indexes of the grid.

The next stage runs in a semi-automatic mode, it determines intrinsic and
extrinsic (relative to the composite 8-sensor system) parameters of the camera
using \gls{lma}.

\begin{figure}[t]
\begin{center}
\includegraphics[width=1.0\linewidth]{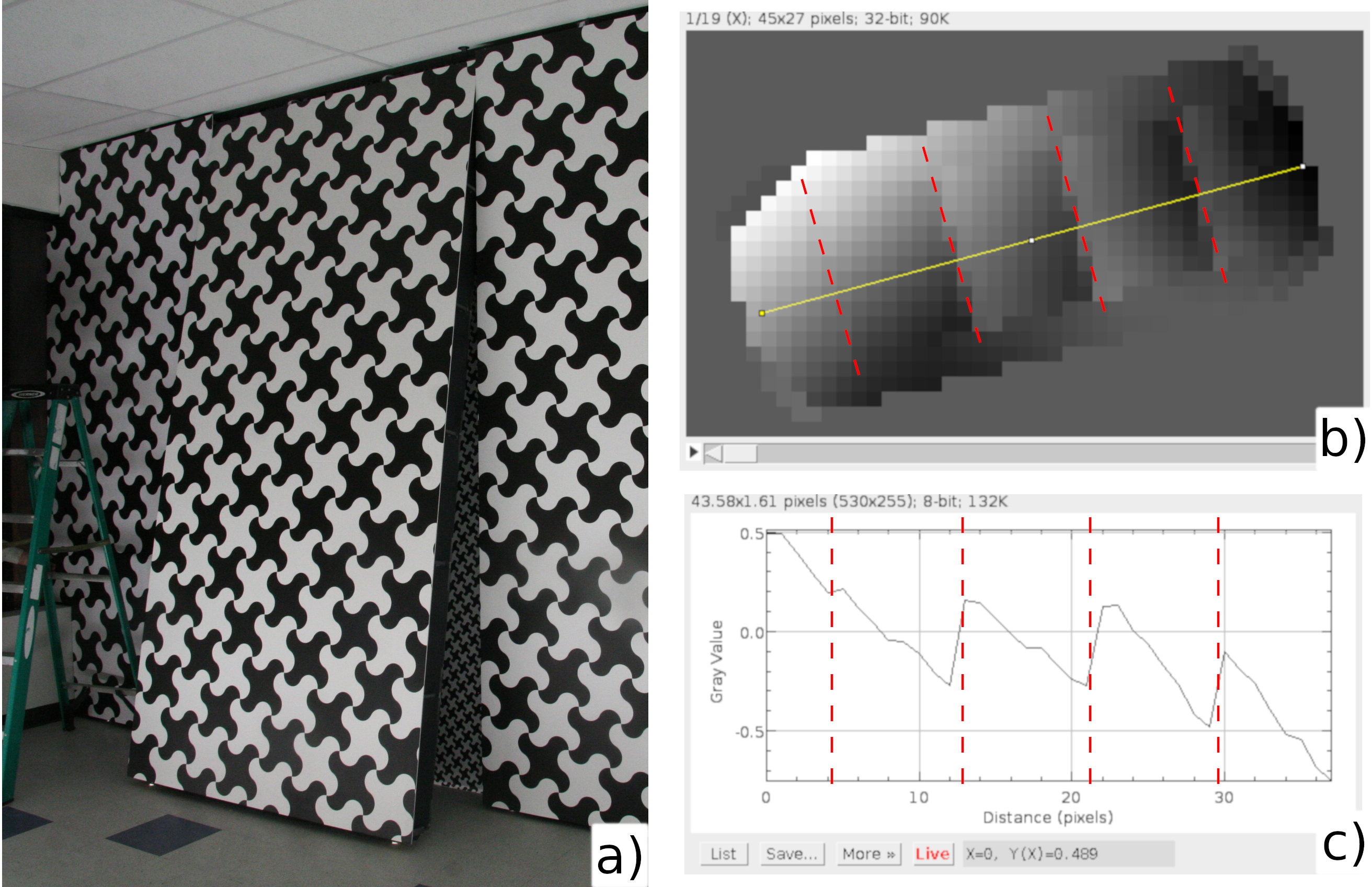}
\end{center}
\caption{Refining pattern nodes coordinates: a) Mounting of the
calibration pattern panels; b) Measured deviation of the nodes in the X
direction (pixel values), tiles correspond to integer grid nodes indexes; c) Profile
along the half-height of the pattern (vertical units are millimeters).}
\label{fig:pattern_tiles}
\end{figure}

First, only visible range cameras data is used to determine parameters of the
high resolution visible range subsystem and simultaneously determine the pose of
the system as a whole. Mechanical model of the camera rotation fixture is
considered: the camera can be tilted \SI{\pm 90}{\degree} and then panned around
its vertical axis. Images are registered from 3 locations, the corresponding
constraints on the extrinsic parameters are imposed. Fitting starts with the
most intrinsic parameters frozen, they are enabled in the process.
Initially, the target is assumed to be ideal: flat and having equidistant grid,
later each grid node location is corrected in 3D space simultaneously with the camera
parameters adjustment.
This is important as the individual panels of the pattern can not be
perfectly matched to each other, and the surface is not flat.
Figure~\ref{fig:pattern_tiles}a) shows mounting of the pattern panels (four
of \SI{1.5 x 3.0}{\meter} and the leftmost \SI{1.0 x 3.0}{\meter}),
\ref{fig:pattern_tiles}b) represents per-node in-plane horizontal correction.
The image is tilted by \SI{15}{\degree} angle corresponding to the pattern
orientation. Red dashed lines indicate seams between the panels, and the yellow
line represents pattern half-height -- a profile line of the
\ref{fig:pattern_tiles}c). Steps of approximately \SI{0.5}{\milli\meter} are
caused by the corresponding gaps between the mounted panels.

When the composite camera extrinsic parameters are determined from the visible
range quadocular subsystem, intrinsic and extrinsic (relative to the composite
camera) parameters of the \gls{lwir} modules are calculated.

The third calibration stage is fully automatic again. Here the space-variant
\glspl{psf} are calculated for each sensor, each color channel and each of the
overlapping by 50\% \SI{256 x 256}{} area of the RGB and \SI{32 x 32}{} area of
the \gls{lwir} image. These \glspl{psf} are subsequently inverted and used as
deconvolution kernels to compensate optical aberrations of the lenses. The
\gls{psf} kernels are calculated by co-processing of the acquired images and the synthetic
ones generated by applying intrinsic and extrinsic parameters, and
calibration pattern geometric corrections calculated in stage 2 to the ideal
pattern model. In addition to accounting for the optical aberrations such as
spherical or chromatic (for RGB modality), these kernels absorb deviations
of the actual lens distortion from the radial model described by each
sub-camera intrinsic parameters. This is especially important for the
\gls{difrec} proposed by Filippov~\cite{filippov2018method}.

\subsection{Differential Rectification of the Stereo Images}
\label{subsec:differntial_rectification}

While long range stereo depends on high disparity resolution, it does not
need to process image sets with large disparities. There are other efficient algorithms
for 3D reconstruction for shorter ranges, such as semi-global matching by
Hirschmuller~\cite{hirschmuller2005accurate} that has \gls{fpga} and \gls{asic}
implementations. It is possible to combine the proposed long range approach
with other algorithms to handle near objects.

\gls{difrec} is partial rectification of the stereo images to the common for all
stereo set distortion model rather than full rectification to a
rectilinear projection.
When \gls{difrec} is applicable it avoids image
re-sampling that either introduces sampling noise or requires up-sampling
that leads to an increase in memory footprint and computational resources.

Limitations of the \gls{difrec} compared to the full rectification are
estimated below, defining maximal disparity that can be processed as well as
requirements to the camera lenses.

Considering the worst correlation case where patch data is all zero except at the very end ($w/2$ for 1-D correlation), disparity error
$d_{err}$ caused by the scale mismatch $K_{diff}$ between the same object
projection in two cameras is
\begin{equation} \label{eq:d_err}
d_{err} = \frac{K_{diff} \cdot w}{2}
\end{equation}
and so for the specified disparity error
\begin{equation} \label{eq:K_diff}
K_{diff} = \frac{2 \cdot d_{err}}{w}
\end{equation}
for $w = 8pix$ and target error of $0.05pix$ $K_{diff}=1.25\%$.

There are two main sources of the $K_{diff}$ considering that the axes of the
camera modules in a rig are properly aligned during factory calibration.

One of them, $K_{lens}$ depends on the differences between the lenses, primarily
on their focal length and can be estimated as \gls{rsd} of the lens focal length
$f$. In our case for the RGB subsystem $\gls{rsd}_{f} = 0.027\% \ll 1.25\%$,
for the \gls{lwir} subsystem $\gls{rsd}_{f} = 1.7\% > 1.25\%$ and may
limit disparity resolution in some cases. $\gls{rsd}_{f}$ can be significantly
improved in production by measuring and binning lenses to select matching
ones for each rig. In our experiments, we did not have extra \gls{lwir}
modules and had to use available ones. 

Disparity term $K_{disp}$ depends on disparity value and the lens distortion.
For the simple case of the barrel/pincushion distortion specified in percents
$D_{perc}$, sensor \gls{fov} radius $r$ and disparity $d$ in pixels

\begin{equation} \label{eq:d_disp}
K_{disp} = \frac{D_{perc}}{100\%} \cdot \frac{d}{r}
\end{equation}
and so maximal disparity for which \gls{difrec} remains valid
\begin{equation} \label{eq:d_max}
d_{max} =  r \cdot \frac{100\%}{D_{perc}} \cdot K_{disp}
\end{equation}
For the RGB cameras $D_{perc} = 10\%$, $r = 1296~pix$ and so $d_{max}(RGB) =
162~pix$, for \gls{lwir} $D_{perc} = 15\%$, $r = 80~pix$ and so $d_{max}(\gls{lwir}) =
6.7~pix$.

\gls{fov} of the visible range cameras in the experimental rig is 20\% larger
than that of the \gls{lwir} one, and with the $16\times$ lower sensor resolution
one \gls{lwir} pixel corresponds to 12.8 pixels
of the RGB cameras, so maximal disparity of \SI{6.7}{\pixel} corresponds for the
same objects to $86.8~pix < d_{max}(RGB) = 162~pix$ of the RGB cameras.

The above estimation proves that \gls{difrec} is justified for our experimental
system with \gls{lwir} disparities up to \SI{6.7}{\pixel} corresponding to the
minimum object distance of \SI{4.0}{\meter}. Other existing algorithms may be
employed to measure shorter distances, but it is possible to relax the above
limitations of the \gls{difrec} for short distances. At a short distance, larger
disparity errors are often acceptable as they cause smaller depth errors, and
the images themselves will be degraded anyway by the depth of field limitation.

\section{Experimental Setup, Image Sets Acquisition and Processing}
\label{sec:acquisition}

\subsection{Dual Quadocular Camera Rig}
\label{subsec:camera_rig}

The experimental camera setup (Figure~\ref{fig:oleg_talon}) consists of the
rigidly assembled quadocular \SI{2592 x 1936}{} RGB camera and
another quadocular \SI{160 x 120}{} \gls{lwir} one. The rig is mounted on a
backpack frame, it is powered by a 48VDC Li~Po battery providing several hours
of operation.

The whole rig was calibrated as described in
Sections~\ref{subsec:calibration_pattern},~\ref{subsec:calibration_processing},
using the \gls{difrec} method. Common sets of the intrinsic parameters were
calculated separately for each of the two modalities. Residual deviations of the
same modality cameras from the respective distortion models where absorbed
by the 2D arrays of space-variant deconvolution kernels saved as the \gls{fd}
coefficients. Each of the 8 cameras was synchronized. Inter-modality offset was measured
by the filming of the rotating object; the result was added to the color
channels to compensate for the latency of the \gls{lwir} modules. FLIR Lepton~3.5 does not
have provisions for external synchronization so we used the genlock approach
with simultaneous reset and simultaneous radiometric calibration. The measured
mismatch between the 4 modules was less than \SI{10}{\micro\second} and it did not change
over tens of hours cameras were running without a reboot.
Color cameras were configured to run in slave mode triggered by the
\gls{lwir} ones, the whole system was operating at \SI{9}{\hertz} limited by
Lepton modules.

\subsection{Image Sets Acquisition}
\label{subsec:acquisition}

Image sets were recorded in raw mode to cameras internal SSD memory,
14 available bits of the \gls{lwir} data were recorded as Tiff files together
with the provided telemetry metadata. Each image file name contains
master camera timestamp and a channel number -- this
information allows to arranging of the image sets of 4 of RGB and 4 \gls{lwir}
each for later processing. Image sets were captured in the
natural environment and contain trees, rocks, snow patches, people and cars -
parked and passing by at highway speed.
Examples of both modality images are
shown in Figure~\ref{fig:tpnet_lwir}, more are available
online~\cite{afilippov2019tpnet} in an interactive table.

While processing the captured sets we found that the 3D-printed camera cases of
the \gls{lwir} modules yielded under pressure of the tightened screws, and we
perform field calibration of the extrinsic parameters. We captured far mounting
ridges for the RGB modality calibration and used an RGB-derived depth map to
calibrate the \gls{lwir} subsystem.

\subsection{Image Sets Processing}
\label{subsec:processing}

Processing of each modality image quads is performed separately, it is based on
the correlation of the \SI{16 x 16}{} overlapping by 50\% tiles.
Subpixel disparity resolution for matching image patches maybe be influenced by
the pixel-locking effect especially when the number of participating pixels is
small. This effect is  described for \gls{piv} applications by Fincham and
Spedding~\cite{fincham1997low}, Chen and Katz~\cite{chen2005elimination}
proposed a method of reducing this effect for clusters over \SI{4 x 4}{}. Pixel
locking for the stereo disparity may occur even for large patches: Shimizu and
Okutomi~\cite{shimizu2005sub} measured this effect by moving the target away
from the camera and proposed a correction.

We use phase correlation for accurate subpixel matching of the image
patches. \gls{difrec} eliminates sampling errors, the subpixel-accurate
initial alignment of the patches is implemented by integer pixel shift followed
by the transformation to the \gls{fd} and phase rotation that is equivalent to
the pixel domain shift. Phase correlation in the \gls{fd} is proven to be free of 
pixel locking effect, Balci and Foroosh\cite{balci2005inferring}
implemented plane fitting to the \gls{fd} view of the phase correlation.
Hoge\cite{hoge2003subspace} calculated disparity from the phase correlation
using the \gls{svd} approach.

\begin{figure*}
\begin{center}
\includegraphics[width=1.0\linewidth]{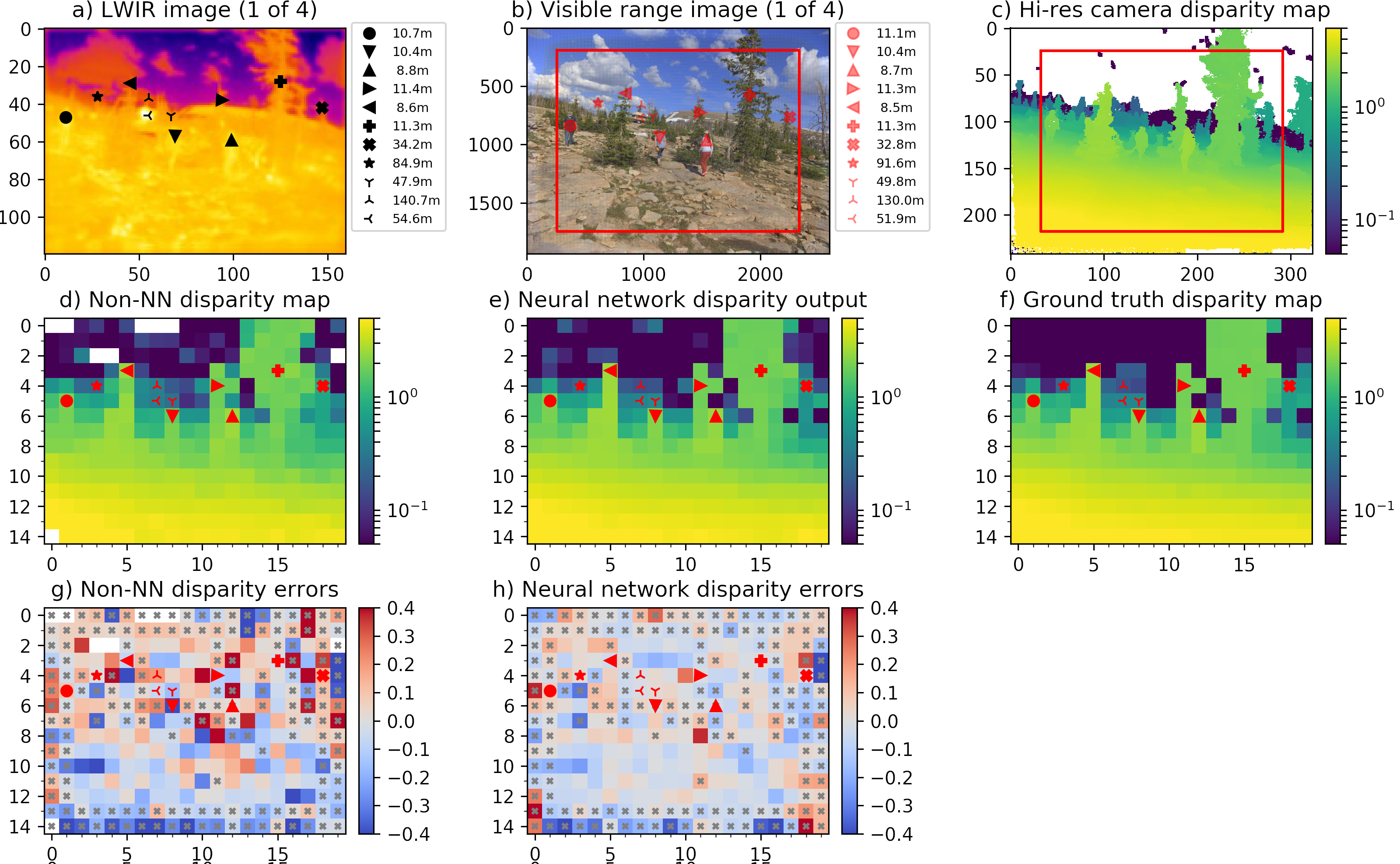}
\end{center}
   \caption{\gls{lwir} depth map generation and comparison to the ground truth
   data for Scene 7 listed in Table~\ref{tab:scenes}.}
\label{fig:tpnet_lwir}
\end{figure*}

The overall method of depth map generation for both modalities involves a fast
converging iterative process where the expected disparity is estimated for each
depth sample, and this value is used for lossless calculation of the \gls{fd}
representation of the corresponding source images patches. This disparity
estimation and corresponding pre-shift of the patches is analogous to the eye
convergence in the human binocular vision. Initial disparity estimation methods
depend on specific application -- expected disparity may be calculated from the
previous sample when a continuous video stream is available.
Other methods use intra-scene neighbors or just simple but time-consuming
disparity sweep. After the residual disparity is calculated from the
correlation, the refined disparity is used in the next iteration step. There are
6 pairs to be correlated in the quadocular system. For the cameras located in the
corners of a square, there are two horizontal pairs, two vertical ones and
two diagonal. For the ground truth depth measurement with color cameras that
have over $13\times$ higher resolution than the \gls{lwir} modality we combine
those six 2D phase correlation results and calculate subpixel disparity by the
polynomial approximation around the integer argmax.
Figure~\ref{fig:tpnet_lwir}b) shows one of the four color images and
\ref{fig:tpnet_lwir}c) illustrates the disparity map calculated from it,
subsequently used to derive the ground truth for \gls{lwir} subsystem
evaluation and training/testing of the \gls{dnn}.

We used the same method as described above to process the 2D correlation outputs
into the disparity map for the \gls{lwir} modality that is the subject of this
research.
Figure~\ref{fig:tpnet_lwir}a) shows registered \gls{lwir} image in
pseudo-colors, and \ref{fig:tpnet_lwir}d) -- corresponding disparity map. Ground
truth for \gls{lwir} shown in \ref{fig:tpnet_lwir}f) is calculated from the
high-res disparity map \ref{fig:tpnet_lwir}c).
As the resolution of the color modality depth map is
higher than that of \gls{lwir}, multiple source tiles map to the same
destination one.
Many destination tiles fall on the edges and get data from both the \gls{fg}
and the \gls{bg} objects.
Averaging disparity values in the destination tiles would result in false
objects at some intermediate between \gls{fg} and \gls{bg} disparity.
Instead, the disparity data for each destination tile is sorted and if
the distribution is found to be bimodal, only the larger disparity values
corresponding to the \gls{fg} object are preserved as the \gls{fg} objects are more
important in most cases.

\subsection{Neural Network Training and Inference}
\label{subsec:nn}

The use of the \glspl{dnn} for the fusion of the multimodal stereo images is now
a popular approach, especially when the application area is defined  in advance,
\eg \gls{dnn} is used for pedestrian detection
(Zhang~\etal~\cite{zhang2019weakly}).
In our work, we do not target any specific scene types but rather aim to improve
subpixel argmax of the 2D correlation, using a minimal amount of prior
information - just that we need long range 3D perception. We adapted TPNET
described by Filippov and Dzhimiev~\cite{filippov2018see} to use with the
\gls{lwir} camera.
That adaptation gave us $2\times$ depth accuracy improvement over the NN-less
method.
Application-specific networks can be built on top of the proposed system.
The network has two stages connected in series:
the first stage consisting of 4 \gls{fc} layers (256-128-32-16) with leaky ReLU
activation for all but the last layer is fed with the correlation data from all
pairs of a single tile together with the amount of applied pre-shift of the
image patches; the second (convolutional with $5\times 5$ kernel, stride~1)
stage receives $20\times 12 \times 16$ output from the first stage and outputs
$20\times 12$ disparity predictions. Such 2-stage architecture, where the first
stage processes individual tiles separately, and the second one adds neighbors
context improves training and reduces computations when only a small fraction of
the input tiles are updated during iteration. The neighbors' context is used
to fill the gaps in the textureless areas and to follow the edges of the \gls{fg}
objects.

For training and corresponding testing, the network was reconfigured to a
25-head Siamese one with 25 instances of Stage~1 fed with \SI{5 x 5}{} patches of
correlation tiles data followed by a Stage~2 instance that outputs a single
disparity prediction corresponding to the center tile of \SI{5 x 5}{} input
group. We used 1100 image sets split as 80\%/20\% for training and testing as a
source of \SI{5 x 5}{} tile clusters. For the cost function, we used the L2 norm
weighted by the ground truth confidence and supplemented it by extra terms to
improve prediction quality and training convergence. Typical 3D scene
reconstructed from the long range stereo images when disparity resolution is
insufficient to resolve depth variations across individual far objects
can be better approximated by a set of fronto-parallel patches corresponding to different
objects than by a smooth surface. The only common exception is the ground
surface that is close to the line of sight, usually horizontal. When just the L2
cost is used the disparity prediction looks blurred. If
the correlated tiles simultaneously contain both \gls{fg} and \gls{bg} features
the result may correspond to the nonexistent object at intermediate range, it
would be more useful if the prediction for such ambiguous tiles would be either
\gls{fg} or \gls{bg}, so we added a cost term for the predictions falling
between the \gls{fg} and the \gls{bg}.

Another cost terms were added to reduce overfitting. Stage~2 prediction for each
tile depends on that tile data and the tiles around it (up to $\pm 2$) in each
direction, and while these other tiles improve prediction by following edges and
filling the low-textured gaps, in most cases just a single tile correlation data
should provide reasonably good prediction, and being a much simpler network it
is less prone to overfitting and plays regularization role when mixed to the
cost function. We added two modified Stage~2 networks with the weights shared
with the original Stage~2 -- one with all but the center Stage~1 output zeroed
out, the second one preserved inner \SI{3x3}{} tile cluster.
The L2 norms from these additional outputs are multiplied by hyperparameters and
added to the cost function, output disparity prediction still uses only the full
\SI{5 x 5}{} Stage~2 outputs.

After the network was trained and tested on the \SI{5 x 5}{} tile clusters
without a larger context we manually selected a 19-scene subset of the
test image sets. These scenes represent different objects (people, trees, rocks, parked
and moving cars), with or without significant motion blur during image
capturing. In addition to the calculation of the \gls{rmse} for all sets, we
marked some features of interest and evaluated the measured distance. \Gls{rmse}
calculated for the whole image would be dominated by the ambiguity in
the attribution of the \gls{fg} tiles to the \gls{bg} (or vice versa) for the
tiles on the object edges, so we removed 10\% outliers from each scene in Table~\ref{tab:scenes}.
This table lists \gls{lwir} disparity errors calculated directly from the
interpolated argmax of combined 2D phase correlation tiles in column 3, the
last column contains errors of the network prediction.
\begin{table}[t]
  \begin{center}
  \begin{tabular}{@{}|c|c|c|c|@{}}
     \hline
    \# &
    Scene timestamp &
	\begin{minipage}[t]{0.21\columnwidth}%
	\centering
	Non-DNN disparity \gls{rmse} (pix)
	\end{minipage} &
	\begin{minipage}[t]{0.21\columnwidth}%
	\centering
	DNN disparity \gls{rmse} (pix)
	\end{minipage}  \\
    \hline
    1  & 1562390202.933097 & 0.136 & 0.060 \\
    2  & 1562390225.269784 & 0.147 & 0.065 \\
	3  & 1562390225.839538 & 0.196 & 0.105 \\
	4  & 1562390243.047919 & 0.136 & 0.060 \\
	5  & 1562390251.025390 & 0.152 & 0.074 \\
	6  & 1562390257.977146 & 0.146 & 0.074 \\
	\textbf{7} & \textbf{1562390260.370347} & \textbf{0.122} & \textbf{0.058 }\\
	8  & 1562390260.940102 & 0.135 & 0.064 \\
	9  & 1562390317.693673 & 0.157 & 0.078 \\
	10 & 1562390318.833313 & 0.136 & 0.065 \\
	11 & 1562390326.354823 & 0.144 & 0.090 \\
	12 & 1562390331.483132 & 0.209 & 0.100 \\
	13 & 1562390333.192523 & 0.153 & 0.067 \\
	14 & 1562390402.254007 & 0.140 & 0.077 \\
	15 & 1562390407.382326 & 0.130 & 0.065 \\
	16 & 1562390409.661607 & 0.113 & 0.063 \\
	17 & 1562390435.873048 & 0.153 & 0.057 \\
	18 & 1562390456.842237 & 0.211 & 0.102 \\
	19 & 1562390460.261151 & 0.201 & 0.140 \\
	\hline
       & Average & 0.154 & 0.077 \\
     \hline
  \end{tabular}
  \end{center}
  \caption{Disparity \gls{rmse} calculated between \gls{lwir} and
  pseudo-ground truth data measured with high resolution visible range
  cameras for 90\%
  of the depth samples. Traditional (non-\gls{dnn}) and \gls{dnn} results are
  presented. Scene 7 details are shown in Figure~\ref{fig:tpnet_lwir}.}
  \label{tab:scenes}
\end{table}

Figure~\ref{fig:tpnet_lwir} illustrates Scene 7 (shown in bold in
Table~\ref{tab:scenes}), other listed scenes with animated views are available
online~\cite{afilippov2019tpnet}.
Scenes 3 and 5 were captured with large horizontal motion blur,
scenes 18 and 19 contain objects closer than the near clipping plane for
disparities.

\section{Discussion}
\label{sec:discussion}

Similar to other researches who work in the area of \gls{lwir} 3D perception
we had to develop a multimodal camera rig to register ground truth
data, develop the calibration pattern, capture and process imagery. There are
multimodal stereo image sets available, such as
Treible~\etal~\cite{treible2017cats} that they subsequently used for WILDCAT network development
Treible~\etal~\cite{treible2019wildcat} and LITIV
(Bilodeau~\etal~\cite{bilodeau2014thermal}) with annotated  human silhouettes.
These benchmark datasets are very useful for evaluation of higher-level
\glspl{dnn}, but they assume specific calibration methods and so are not
suitable for comparison of end-to-end systems that mix hardware, calibration, 
and software components. The ultimate test of the 3D perception system is how
well it performs its task, how reliable is the autonomous driving, but such
tests (Zapf~\etal~\cite{zapf2017perception}) require specific test sites and so
are difficult to reproduce when different hardware is involved.

Comparison of just the calibration methods has its limitations too -- measured
\gls{mre} value may be misleading when not verified by the actual disparity
accuracy in 3D scene reconstruction performed  independently from the
calibration. 
This is why the calibration quality is often evaluated by the accuracy of the
stereo 3D reconstruction, with the result presented as disparity resolution in
pixels -- parameter that is invariant of the sensor resolution, focal length to
pixel size ratio (angular resolution), and the camera baseline.

We follow this path and compare our results with those published in
Lee~\etal~\cite{lee2016lwir}. They used a pair of the FLIR Tau 11 \SI{640x512}{}
cameras at a baseline of \SI{0.75}{\meter} with the ground truth data provided
by a LIDAR. During processing, they used a two-level representation with L0
having full \SI{640x512}{} resolution and L1 -- reduced to \SI{320 x 256}{}.
They compared theoretical models of range accuracy for distances of up to
\SI{40}{\meter} range for \SI{0.25}{\pixel} disparity resolution of L0 and L1
and the measured range accuracy at up to \SI{25}{\meter}, their graphs show
match for L1 with \SI{0.25}{\pixel} resolution, corresponding to
\SI{0.5}{\pixel} of the registered images. Our results are
\SI{0.154}{\pixel} with polynomial argmax interpolation and \SI{0.077}{\pixel}
with a trained \gls{dnn} -- 6.5 times improvement.
While our system uses lower resolution \gls{lwir} sensors, the dual-modal
calibration method, and quadocular camera design demonstrate that disparity
accuracy in pixels remains approximately constant even for much higher
resolution visible range cameras.

Comparison of our disparity density results with
their published disparity map examples (63\%-70.5\%) shows our system
advantage (Table~\ref{tab:scenes} summarizes 90\% of the best tiles), but such
comparison is less strict as the density is highly dependent on the scene
details.

\section{Acknowledgments}
\label{sec:acknowledgments}
We thank Tolga Tasdizen for his suggestions on the network
architecture and implementation. This work is funded by SBIR Contract
FA8652-19-P-WI19 (topic AF191-010).

\setglossarysection{section}

{\small
\bibliographystyle{ieee_fullname}
\bibliography{elphel_lwir3d}
}

\end{document}